\def\year{2022}\relax

\documentclass[letterpaper]{article} 
\usepackage{aaai22} 
\usepackage{times} 
\usepackage{helvet}
\usepackage{amsmath, amssymb}
\usepackage{courier} 
\usepackage[hyphens]{url} 
\usepackage{graphicx}
\usepackage{lipsum}
\usepackage{xcolor}
\usepackage{natbib} 
\usepackage{caption} 
\DeclareCaptionStyle{ruled}{labelfont=normalfont,labelsep=colon,strut=off} 
\usepackage{algorithm}
\usepackage{algpseudocode}

\usepackage{newfloat}
\usepackage{listings}
\usepackage{booktabs}
\usepackage{subcaption}
\usepackage{hyperref}
\floatstyle{ruled}
\newfloat{listing}{tb}{lst}{}
\floatname{listing}{Listing}

\nocopyright

\title{On Causal Inference for Data-free Structured Pruning}
\iftrue
\author{
    Martin Ferianc\textsuperscript{\rm 1}, 
    Anush Sankaran\textsuperscript{\rm 2}, Olivier Mastropietro\textsuperscript{\rm 3}, Ehsan Saboori\textsuperscript{\rm 3}, and Quentin Cappart\textsuperscript{\rm 4}
}
\affiliations{
    \textsuperscript{\rm 1}Department of Electronic and Electrical Engineering,
    University College London, London, UK WC1E 7JE, martin.ferianc.19@ucl.ac.uk \\
    \textsuperscript{\rm 2} Independent Researcher, anush.sankaran@gmail.com \\
    \textsuperscript{\rm 3} Deeplite Inc., Montreal, QC, Canada H3C 2G9, \{olivier, ehsan\}@deeplite.ai \\
    \textsuperscript{\rm 4} Department of Computer Engineering and Software Engineering, Polytechnique Montréal, Montreal, QC, Canada H3T 1J4, quentin.cappart@polymtl.ca
}
\fi

\begin{document}

\maketitle

\begin{abstract}
Neural networks (NNs) are making a large impact both on research and industry. Nevertheless, as NNs' accuracy increases, it is followed by an expansion in their size, required number of compute operations and energy consumption. Increase in resource consumption results in NNs' reduced adoption rate and real-world deployment impracticality. Therefore, NNs need to be compressed to make them available to a wider audience and at the same time decrease their runtime costs. In this work, we approach this challenge from a causal inference perspective, and we propose a scoring mechanism to facilitate structured pruning of NNs. The approach is based on measuring mutual information under a maximum entropy perturbation, sequentially propagated through the NN. We demonstrate the method's performance on two datasets and various NNs' sizes, and we show that our approach achieves competitive performance under challenging conditions.

\end{abstract}
\section{Introduction}\label{sec:introduction}

Neural networks (NNs) have been successfully deployed in several applications, such as computer vision~\cite{zhai2021scaling} or natural language processing~\cite{brown2020language}. The accuracy in these tasks increases with ongoing development, however, so does the model size and power consumption~\cite{guo2019empirical}. Novel NNs' increasing size, complexity, and energy demands limit their deployment to low-cost compute platforms.

On one hand, hardware optimizations have been proposed to ease the deployment of demanding NNs, but these are usually targeting on a specific pair of NN architecture and a hardware platform~\cite{chen2020survey}. On the other hand, software optimizations, such as structured pruning~\cite{hoefler2021sparsity} have been proposed, which can be applied to a variety of NNs to make them smaller. In structured pruning, a certain neuron in the NN is removed completely, saving computation time, reducing their memory, and energy consumption across multiple hardware platforms. For example, by compression and subsequent fine-tuning, ResNet-18's compute operations' count can be reduced by $7\times$ and its memory foot-print by $4.5\times$~\cite{sankaran2021deeplite}. However, most of the methods for structured pruning are based on some heuristics which are not conclusive in the context of structurally examining NNs. Additionally, pruning might require access to the original data on which the NN was trained, for further fine-tuning. 

To address these challenges, in this research work we propose first steps towards a data-free approach to structured pruning which is facilitated through causal inference. In this approach, we evaluate the importance of neurons by measuring mutual information (MI) under a maximum entropy perturbation (MEP) propagated through the NN. We demonstrate our performance and generalizability on various fully-connected NN architectures on two datasets. In our evaluation, the proposed method produces marginal improvements over the existing work and it hopes to pave new directions for research into causal inference applied to optimize NNs.

\section{Related Work}\label{sec:related_work}

\subsection{Causal Inference and Information Bottleneck}

Our method is primarily inspired by the work of~\citet{mattsson2020examining} who have proposed a suite of metrics based on information theory; to quantify and track changes in the causal structure of NNs. They introduced, the notion of \textit{effective information} which is the MI between layer input and output following a local layer-wise MEP. We build on this notion and introduce several changes. First, we sample a random intervention only at the input of the NN and we measure the MI with respect to the output of the previous layer obtained by propagating the intervention throughout the net. Second, we pick a different maximum entropy distribution, a Gaussian instead of uniform, that more closely reflects real-world data. Third, we combine the different measurements per neural connection, and we use them to score the likeliness of that neuron for structured pruning. Additional concept related to this work is \textit{information bottleneck}~\cite{tishby2000information}, which measures MI with respect to the information plane and propagating data through the network. They have shown that at a certain point in the NN, the NN minimizes MI between input and output. In this work, we contemplate that if Gaussian noise is propagated through the net, the neurons which maximize the MI between input and output are preferred with respect to generalization on the test data.

\subsection{Structured Data-free Pruning}

\citet{hoefler2021sparsity} completed a comprehensive survey of NN pruning methods and in this work we will focus on those that: do not require data to prune and focus on structured pruning. \citet{srinivas2015data} proposed a \textit{data-free pruning} (DFP) method that examines the importance of different neurons based on their similarity through the magnitude of their weights. Their method iteratively examines, prunes and updates this similarity along with the weights of the NN. \citet{mussay2019data} proposed a data-independent way for pruning neurons in an NN through coreset approximation in its preceding layers. \cite{wang2019cop} developed \textit{correlation-based pruning} (COP), which can detect the redundant neurons efficiently through removing the ones which are the most correlated with the others. Moreover,~\citet{narendra2018explaining} developed a method to reason over NN as a structured causal model, nevertheless, this method is data-bound. Lastly,~\citet{ganesh2021mint} introduced \textit{MINT} which is based on measuring MI with respect to data, however, without considering the notion of causal inference or MEP. With respect to the related work, our method also wants to appeal to users who seek data-free pruning methods, potentially due to privacy-related constraints. Nevertheless, our method differs by avoiding the usage of heuristics, such as: the weight magnitude or correlation. It relies on examining the causal structure in the NN, rather than deterministic heuristics. 
\section{Method}\label{sec:method}

Without using the train data, NNs and their internal connectivity have been often described through heuristics, such as correlations and magnitude of the connecting weights for the individual neurons~\cite{hoefler2021sparsity}. As the depth and width of the NNs increase, these metrics become less transparent and less interpretable in feature space. Additionally, there is no clear link between these heuristics and the causal structure by which the NN makes decisions and generalizes beyond train data. As it has already been argued, generalizability must be a function of a NN’s causal structure since it reflects how the trained NN responds to unseen or even not-yet-defined future inputs~\cite{mattsson2020examining}. Therefore, from a causal perspective, the neurons which are identified to be more important in the architecture should be preserved and the ones that are identified less important could be removed. This paradigm paves the way for observing the causal structure, identifying important connections and subsequent structured pruning in NNs, replacing heuristics, to achieve better generalization.

\begin{algorithm}
\caption{Inference of MI scores for structured pruning.}\label{alg:main_algorithm}
\begin{algorithmic}[1]
\State \textbf{\textit{Phase 1: Record and normalize inputs and outputs}}
\State Sample $S$ samples for NN input $\boldsymbol{x}_0{\sim}\mathcal{N}(0,1)$ 
\State Pass $\boldsymbol{x}_0$ and cache layers' outputs $\boldsymbol{x}_i;i\in [1,L]$
\State {Infer range of activations for each layer $i\in [0,L]$} 
\State {Clamp and normalize each $\boldsymbol{x}_i; i\in [0,L]$ }
\State \textbf{\textit{Phase 2: Infer input-output mutual information}}
\State Initialize MI as an empty list
\For{$(\boldsymbol{x}_{i-1}\in\mathbb{R}^{S\times N}, \boldsymbol{x}_{i}\in\mathbb{R}^{S\times M})$;$i\in [1,L]$}
    \State Initialize $\boldsymbol{mi}_i\in\mathbb{R}^{N\times M}$ as zeros
    \For{$n$ in range $N$}
        \For{$m$ in range $M$}
             \State { $\boldsymbol{h}_{i}^{n,m} =$ joint histogram for $\boldsymbol{x}_{i-1}^n,\boldsymbol{x}_{i}^m$}
            \State $mi_i[n,m]=$ Mutual information for $\boldsymbol{h}_{i}^{n,m}$
        \EndFor
    \EndFor
    \State Add $\boldsymbol{mi}_i$ to MI to be used during structured pruning 
\EndFor
\State \textbf{\textit{Phase 3: Pruning}} 
\For{layer $i; i\in [0, L-1]$}
    \State $\boldsymbol{mi}$ = MI[i]
    \State Set $\boldsymbol{mi}$ indices of already pruned neurons to zero
    \State \textit{Scores} = Sum $\boldsymbol{mi}$ row-wise
    \State Sort \textit{Scores}
    \State Remove neuron connections for lowest \textit{Scores}
\EndFor
\end{algorithmic}
\end{algorithm}

In this work, we propose a perturbation-based approach to study the causal structure of the NN which enables us to quantify the significance of each neuron in the NN. The method performs an intervention $do(\boldsymbol{x}_0)$ at the input level of the NN. However, instead of choosing a single type of intervention we opt for a maximum entropy distribution - a Gaussian distribution, which covers the space of all potential interventions with a fixed variance, and it is used to sample the input $\boldsymbol{x}_0{\sim}\mathcal{N}(0,1)$.  Subsequently, the output of the input layer is propagated to deeper layers $i;i\in1,\ldots,L$ up to the entire depth $L$ of the net. Under this perturbation, we propose to measure MI between the input and output pairs of layers, to measure the strength of their causal interactions~\cite{mattsson2020examining}. Unlike the standard MI, which is a measure of correlation, all mutual bits with a noise injection will be caused by that noise~\cite{mattsson2020examining}. We hypothesize that the connections that preserve the most information on average under MEP are the strongest, with the most impact on the generalization performance, and they should be preserved in case of pruning. The MI is measured individually per input-output connection, and later summed for the given output neuron, for computational simplicity. However, the individual computation and summation imply independence with respect to the input connections for a particular neuron. This is a limitation that manifests itself with increasing depth of the net.

\begin{figure*}[t]
    \centering
    \includegraphics[width=1\linewidth]{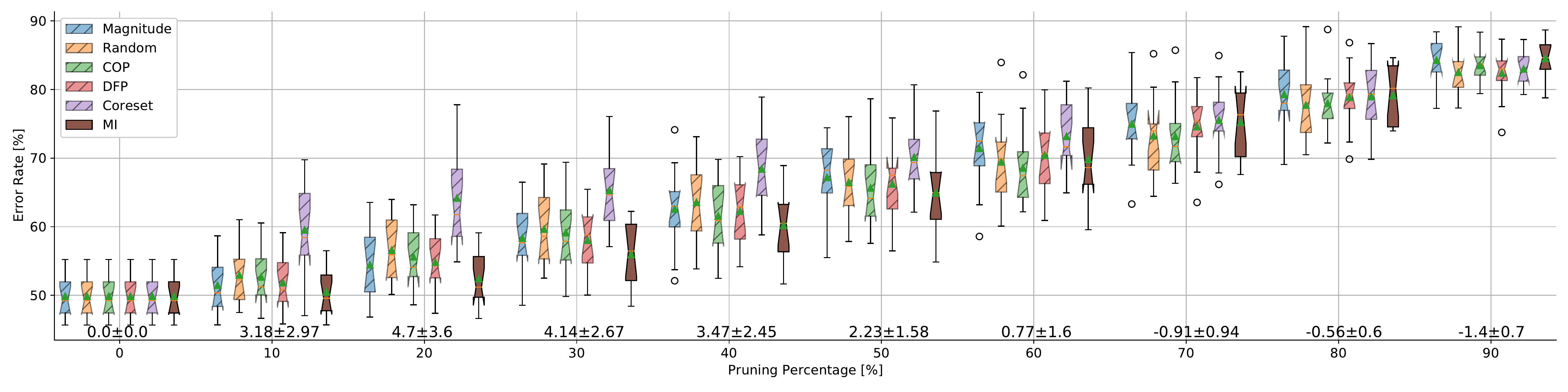}
    \vspace{-1em}
    \caption{CIFAR-10 test dataset results, each box is aggregation of all 12 networks pruned with respect to the set percentage.}
    \label{fig:cifar10_box_test_error}
\end{figure*}
\begin{figure*}[t]
    \centering
    \includegraphics[width=1\linewidth]{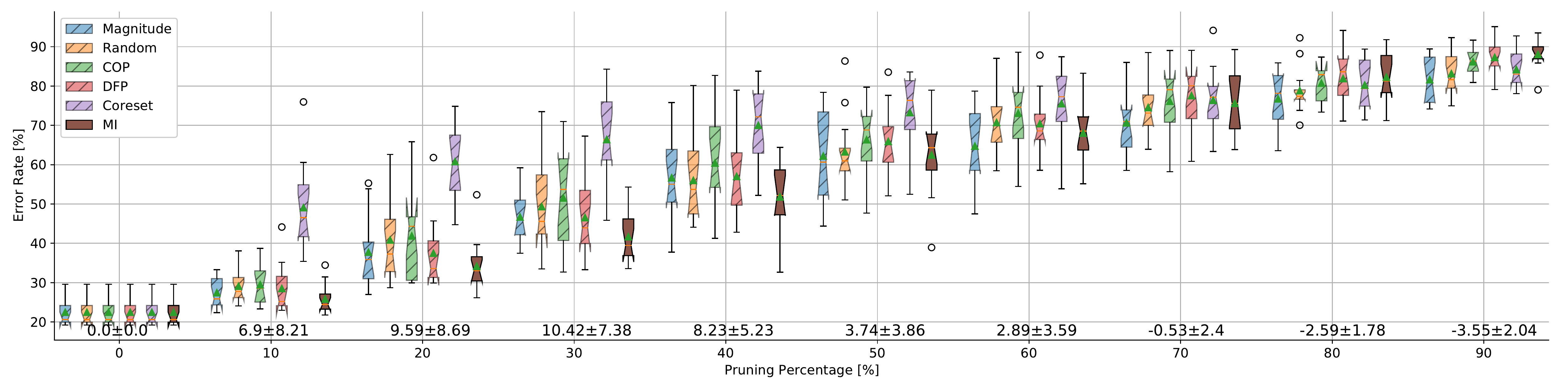}
    \vspace{-1em}
    \caption{SVHN test dataset results, each box is aggregation of all 12 networks pruned with respect to the set percentage.}
    \label{fig:svhn_box_test_error}
\end{figure*}

The Algorithm~\ref{alg:main_algorithm} summarizes the method. In general, it consists of three phases: 1. recording and normalization of NN's layers' inputs and outputs, 2. inference of MI from the records and 3. using the MI to determine which neurons to prune. The first phase starts with propagating the random noise through the trained NN, while caching, clamping and normalizing the outputs of all layers between $[0,1]$, for the inferred maximums and minimums of their paired activations. In practice, the normalization was used to standardize the relative bin size of the input-output histograms that are being used for the MI estimation. 

Given the cached inputs and outputs for all layers $i\in 1,\ldots,L$, the algorithm then proceeds to the second phase to estimate per-connection MI for each layer $i$. For example, if considering an input from the previous layer $\boldsymbol{x}_{i-1}\in \mathbb{R}^{S\times N}$ with $N$ nodes and $S$ total samples and the output of the current layer, after an activation was applied, $\boldsymbol{x}_{i}\in \mathbb{R}^{S\times M}$ with $M$ output nodes, we calculate the MI for a connection between $n\textsuperscript{th}$ and $m\textsuperscript{th}$ neuron and the $i\textsuperscript{th}$ layer as $mi_{i}^{n,m}(\boldsymbol{x}_{i-1}^n,\boldsymbol{x}_{i}^m | do(\boldsymbol{x}_0));\boldsymbol{x}_0{\sim}\mathcal{N}(0,1)$, where $mi$ stands for MI estimation based on a histogram~\cite{fraser1986independent}.
In detail, the MI for node pair $n,m$ and layer $i$ as $mi_{i}^{n,m}$ is computed as shown in Equation~\ref{eq:mi}:
\vspace{-0.5em}
\begin{multline}
    mi_{i}^{n,m}(\boldsymbol{x}_{i-1}^n, \boldsymbol{x}_{i}^{m}| do(\boldsymbol{x}_0)) = \sum_{u\in \boldsymbol{x}_{i-1}^n} \sum_{v \in \boldsymbol{x}_{i}^{m}} p_{x_{i-1}^n, x_{i}^{m}}(u,v) \\\log \left(\frac{p_{x_{i-1}^n, x_{i}^{m}}(u,v)}{p_{x_{i-1}^n}(u) p_{x_{i}^{m}}(v)}\right)
    \label{eq:mi}
\end{multline}
The different probability densities $p(.)$ are captured by histograms with $B\times B$ bins with respect to their 2D space, defined by $S$ normalized and cached samples of $\boldsymbol{x}_{i-1}^n,\boldsymbol{x}_{i}^m$ for the given input intervention $do(\boldsymbol{x}_0)$. This process is repeated with respect to all $n\in N$ for a particular $m\textsuperscript{th}$ node out of $M$ nodes and for every layer $i\in 1,\ldots,L$ with the same $\boldsymbol{x}_0$. These scores are pre-computed and cached until the pruning phase. 

In the pruning phase, the final significance score for layer $i$ and some node $m$ is determined by summation of the pre-recorded $mi_{i}^{n,m}$ with respect to all existing connections $n\in N$, where the $n\textsuperscript{th}$ link was not cancelled by the pruning in the previous layer. The scores for all $m\in M$ for a given layer $i$ are then sorted and the neurons with the smallest score are pruned, moving to the next layer. Hence, the overall algorithm focuses on preserving the neurons that were observed with the highest input-output MI under the MEP and we hypothesize that they should have the most impact on the generalization performance of the NN.

\section{Experiments}\label{sec:experiments}

To validate the proposed method, we performed comprehensive experiments involving two datasets and various network depths and widths. All networks were trained with respect to 200 epochs, learning rate set initially to $1e^{-3}$ and exponentially decayed with an Adam optimizer and weight decay set to $1e^{-4}$. We conducted experiments with respect to CIFAR-10 and SVHN, to vary the complexity of the datasets, without any data augmentations except normalization. For both datasets we trained in total 12 networks, paired with ReLU activations, with $\{1,2,3\}$ hidden layers and $\{64, 128, 192, 256\}$ channels for CIFAR-10 or $\{16, 32, 48, 64\}$ channels for SVHN, giving 12 model combinations for each dataset. The models were arguably small, where it can be assumed that each neuron has certain importance and there are no or few inactive neurons. Therefore, the pruning methods need to be careful about scoring the neurons, since removing even a single neuron will affect the algorithmic performance. In terms of pruning, we ask each compared method: magnitude-based~\cite{he2017channel}, Random, COP, DFP, coreset or ours (MI) to provide a relative importance score for all hidden neurons in an NN. We used publicly available implementations of the respective methods, except DFP which we reimplemented. We adopted a linearly increasing pruning schedule with respect to depth of a layer with some maximum percentage, omitting the input or output layers. For example, if we set the pruning rate to 30\% and the network has 2 hidden layers, we would prune 15\% of neurons in the 1$\textsuperscript{st}$ hidden layer and 30\% in the 2$\textsuperscript{nd}$ hidden layer depending on the lowest scores given by each method. We used $S=5000$ samples for MI estimation with $B=32$. 

\subsection{Aggregated Results}

In Figures~\ref{fig:cifar10_box_test_error} and~\ref{fig:svhn_box_test_error} we demonstrate the results showing varying error rate across different limiting pruning percentages. Each box represents aggregated results from 12 benchmarked models pruned with respect to the limiting percentage. We chose this form of presentation to demonstrate the versatility of our method and related work across different network depths and widths. As it can be seen with respect to both datasets, our method's error rates increase less in comparison to the related work across a range of different architectures and pruning percentages, especially in lower pruning percentages. Quantitatively, the benefits of our method are shown at the bottom of the plots. The numbers symbolize the mean and the standard deviation of reduction in average error rate across the 12 architectures and the related work compared to the proposed method. Note that, the evaluated networks were small and every connection can be hypothetically significant to generalization and the error increases irrespective of selected method after setting the bounding percentage to approximately 70\%.
\begin{figure}[t]
    \centering
    \includegraphics[width=1\linewidth]{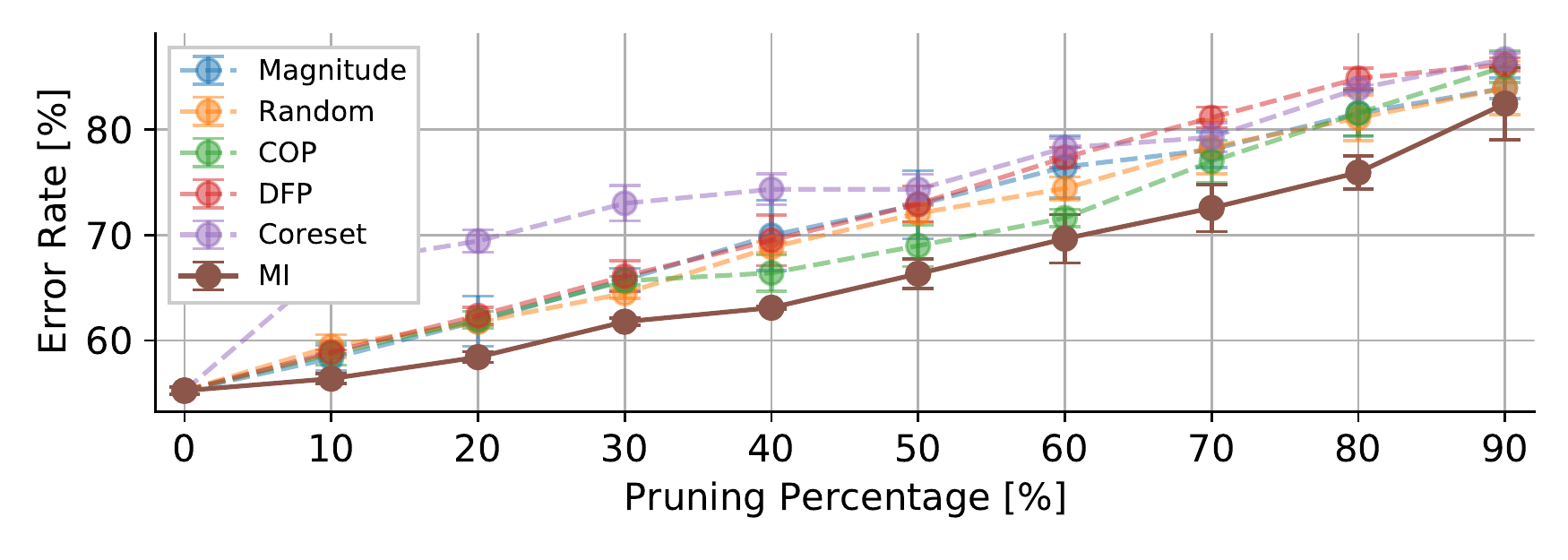}
    \vspace{-2.2em}
    \caption{Network with one hidden layer with 64 channels for CIFAR-10.}
    \label{fig:cifar10_small}
\end{figure}
\begin{figure}[t]
    \centering
    \includegraphics[width=1\linewidth]{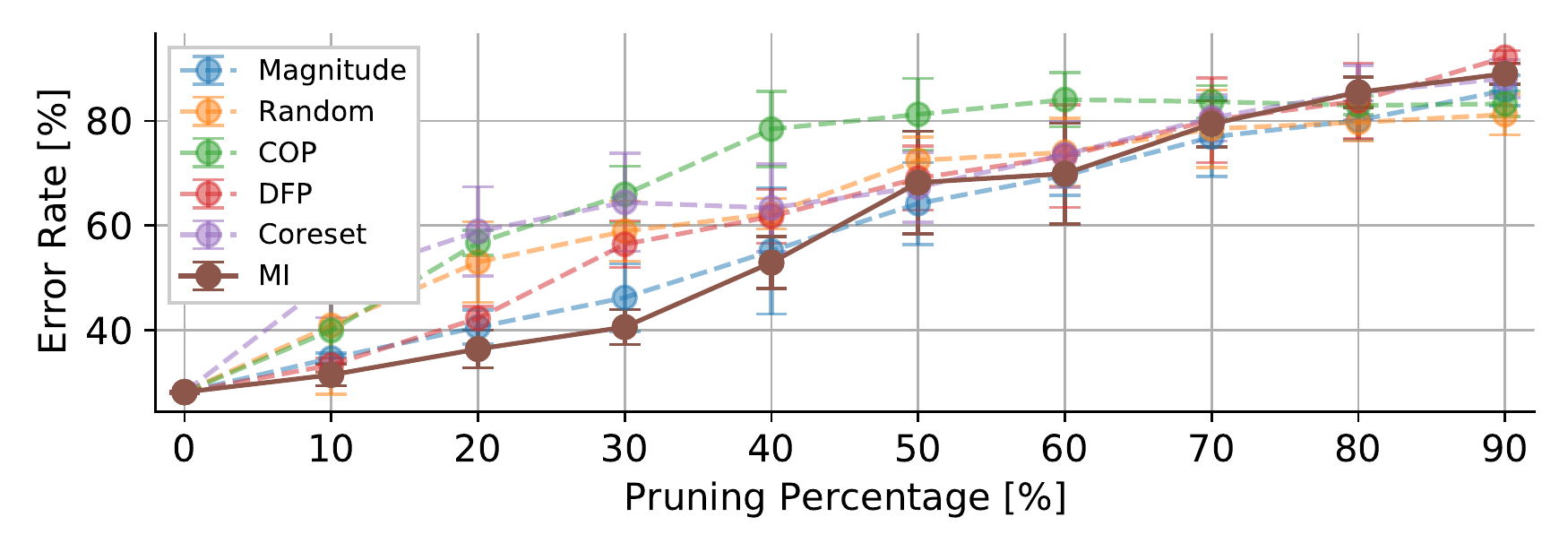}
    \vspace{-2.2em}
    \caption{Network with one hidden layer with 16 channels for SVHN.}
    \label{fig:svhn_small}
\end{figure}
\begin{table}[t]
    \scalebox{0.78}{
    \begin{tabular}{c|c|c|c|c}
    \toprule
       & \multicolumn{2}{c|}{\bf CIFAR-10} & \multicolumn{2}{c}{\bf SVHN}  \\ \cmidrule{2-5}
         & Correlation & Kendall $\tau$ & Correlation & Kendall $\tau$   \\ \midrule
    \textbf{Layer 1}   & $0.86\pm 0.006$ &$0.78\pm 0.01$ & $0.5\pm 0.06$ & $0.3\pm 0.05$   \\ \midrule
    \textbf{Layer 2}   & $0.5\pm 0.1$ & $-0.15\pm 0.04$ & $-0.29\pm 0.05$ & $-0.2\pm 0.03$ \\ \midrule
    \textbf{Layer 3}   & $0.6\pm 0.06$& $0.02\pm 0.04$& $-0.08\pm 0.68$ & $-0.3\pm 0.16$ \\ \midrule
    \textbf{Layer 4}   & $0.9\pm 0.0$& $0.47\pm 0.02$ & $0.74\pm 0.1$ & $0.05\pm 0.1$\\ 
    \bottomrule
    \end{tabular}}
    \caption{Ranking similarity to magnitude-based score for the deepest and widest network variants.}
    \label{tab:large}
\end{table}
\begin{table}[t]
    \scalebox{0.78}{
    \begin{tabular}{c|c|c|c|c}
    \toprule
       & \multicolumn{2}{c|}{\bf CIFAR-10}& \multicolumn{2}{c}{\bf SVHN}  \\ \cmidrule{2-5}
         & Correlation & Kendall $\tau$ & Correlation & Kendall $\tau$   \\ \midrule
    \textbf{Layer 1}  & $0.48\pm 0.07$ & $0.32\pm 0.05$ & $0.11\pm 0.08$ & $0.06\pm 0.12$   \\ \midrule
    \textbf{Layer 2}   & $-0.43\pm 0.02$& $-0.23\pm 0.05$& $-0.01\pm 0.43$ & $0.05\pm 0.42$ \\ 
    \bottomrule
    \end{tabular}}
    \caption{Ranking similarity to magnitude-based score for the shallowest and thinnest network variants.}
    \label{tab:small}
\end{table}
\subsection{Detailed Results}

In Figures~\ref{fig:cifar10_small} and~\ref{fig:svhn_small} we present the results with respect to the smallest and most challenging architectures in our experiments with only one hidden layer. All experiments were repeated 3 times with different random seeds to observe mean and standard deviation for robustness. As it can be seen, MI was able to more concretely identify the significant neurons, resulting in lower average error rates, mainly for CIFAR-10. 

Additionally, in Tables~\ref{tab:large} and~\ref{tab:small} we demonstrate the Spearman correlation and { Kendall $\tau$}
ranking correlation with respect to magnitude-based pruning, which is a well-established baseline, to provide deeper insight into the proposed method. As it can be seen, the method is partially correlated to the magnitude of the weights connecting that neuron to the rest of the NN. However, looking simultaneously at the {Kendall $\tau$} comparing weight magnitude and our score, it can be seen that the overall ranking is completely different. These results demonstrate that causal inference and MI in general are vital for deeper understanding of the structure of the NN and there is only a relatively weak connection to the weights' magnitude. 

\subsection{Challenging Settings}
During the experimentation we made several observations and we also encountered challenging experimental and deployment settings. In initial experiments with respect to MNIST and FashionMNIST, we noticed that the proposed method underperformed. We associated this with respect to atypicality of those datasets, containing predominately zeros resulting in skewed input data distributions, far from the Gaussian assumption. Moreover, we noticed that the performance of the method degrades with increasing the depth of the network. We associated this with respect to the MI estimation, where the independence assumption does not hold and thus MI is extremely challenging to estimate.

\section{Conclusion}\label{sec:conclusion}

In this work, we presented empirical first steps towards a causal inference-based approach for data-free structured NN pruning. We evaluated the proposed methodology with respect to different NN structures on two real-world datasets. Additionally, we detailed positive cases for pruning as well as challenging conditions. In the future work, we aim to extend the current framework with respect to complex networks, specifically convolutional NNs, to promote its practicality in real-world applications. Moreover, we want to investigate the methodology with respect to additional tasks or larger datasets.

\iftrue
\section{Acknowledgements}
This work was completed, while Martin Ferianc was an intern and Anush Sankaran was a research scientist at Deeplite Inc. This research was supported by MITACS/IT26487. This funding source had no role in the design of this study and did not have any role during its execution, analyses, interpretation of the data, or decision to submit results. Lastly, we thank ITCI'22 reviewers for feedback that helped us to improve the paper.
\fi 
\bibliography{main}
\end{document}